# ClimateChat-300K: A Multi-Modal Facebook Dataset for Understanding Diverse Perspectives in Climate Communication

**Wajdi Zaghouani**[1], **Md. Rafiul Biswas**[2], **Mabrouka Bessghaier,**[1]
**Shimaa Ibrahim**[1], **George Mikros**[2]
[1]Northwestern University in Qatar
[2]Hamad Bin Khalifa University, Qatar
{wajdi.zaghouani, mabrouka.bessghaier, shimaa.ibrahim}@northwestern.edu
{mbiswas, gmikros}@hbku.edu.qa

**Abstract**

We present *ClimateChat-300K*, a large-scale dataset of 299,329 public Facebook posts about climate change collected between May 2020 and May 2024 through the CrowdTangle platform. The dataset contains 41 metadata features including post content, engagement metrics, and page attributes, covering material from more than 26,000 global pages. Each post includes rich contextual information such as language, timestamp, page category, and interaction counts, enabling comprehensive analyses of public discourse around climate communication. Using topic modeling and sentiment analysis, we identify ten main themes grouped into five domains: policy, activism, cooperation, science, and conservation. The results reveal that emotional tone, post format, and page identity strongly influence audience engagement, with visually rich and emotionally charged content receiving the highest levels of interaction. The dataset also demonstrates how online discussions evolved in response to major events such as international climate summits and the COVID-19 pandemic period. ClimateChat-300K provides an open resource for reproducible and interdisciplinary research on polarization, misinformation, and the dynamics of digital climate discourse. By releasing this dataset, we aim to support transparent, data-driven research and contribute to a deeper understanding of how public engagement with climate issues develops across time, geography, and institutional contexts.

## 1. Introduction

Climate change is widely regarded as one of the most significant global challenges of the 21st century (Sultana et al., 2024). It threatens the future of communities, ecosystems, and entire generations, demanding an extraordinary level of global cooperation. Scientists overwhelmingly agree that human activities are driving the accelerating changes in our climate (Storani et al., 2025). Yet, despite this clear consensus, public debate remains intense often shaped more by politics and social divisions than by science itself (Deo and Prasad, 2020). Overcoming this crisis calls for more than innovation in labs or new technologies; it requires understanding how people think, talk, and act when it comes to the planet's future.

In recent years, social media has become a central platform for communication and debate surrounding climate change. These digital spaces play a crucial role in shaping public discourse by enabling rapid information exchange, public engagement, and grassroots activism (Sultana et al., 2024; Mede and Schroeder, 2024). Individuals and organizations now use platforms such as Facebook, X (formerly Twitter), and Instagram to coordinate climate campaigns, raise awareness, and mobilize communities globally (Mavrodieva et al., 2019). During 2020-2024, social media platforms witnessed significant shifts in the dynamics of online climate conversations, particularly surrounding major global events. The COVID-19 pandemic forced climate activism's "digitalization," as movements from street protests to online campaigns Springer (Sorce and Dumitrica, 2023), while major climate negotiations such as the COP26 in Glasgow (2021), COP27 in Egypt (2022), and COP28 in Dubai (2023) each generated millions of social media interactions with distinct discourse patterns (Falkenberg et al., 2022; Sultana et al., 2024). Given the influential role of social media in the climate change debate, analyzing social media data offers valuable insights into public behavior and opinion (Segerberg, 2017). Despite the importance of Facebook and other social platforms in climate change communication, there is a notable gap in accessible data and research focus for certain platforms (mostly skewed toward Twitter) (Souza et al., 2014; Segerberg, 2017). Far fewer studies have examined climate conversations on Facebook (Falkenberg et al., 2022) in part because Facebook's data are more difficult to obtain due to privacy restrictions and API limitations (Deo and Prasad, 2020). This imbalance means that our scientific understanding of online climate dialogue may be incomplete or biased toward the Twitter user demographic (Mooseder et al., 2023).

To address this gap, in this paper we present a new dataset of 299,329 Facebook posts with 41 metadata about climate change, collected from a global English-language corpus spanning four years (17 May 2020 to 16 May 2024). By curating and releasing this dataset, we aim to provide the research community with a rich resource to study climate change discourse on Facebook – a platform that remains one of the largest social networks worldwide. The dataset captures a critical period in recent history, including the COVID-19 pandemic (which intertwined with environmental issues) and the ensuing years of intensified climate policy dialogue and activism. Researchers can use this data to perform longitudinal analyses of climate communication, compare discourse patterns between Facebook and other platforms, and explore how public engagement with climate topics has evolved in response to external events or social trends. Moreover, making this dataset openly available helps to overcome some of the data access barriers that have traditionally limited computational social science research on Facebook (Deo and Prasad, 2020; Storani et al., 2025).

## 2. Methods

### 2.1. Data Collection

**Data Source:** We collected data from Facebook Pages using the CrowdTangle platform (Center, 2024), a public insights tool owned by Meta that provided programmatic access to public content from Pages, Groups, and verified accounts. CrowdTangle was widely adopted by journalists and researchers for transparency and social media monitoring until its closure in 2024. As a result, the present dataset constitutes a timely and valuable resource, preserving access to climate-related communication that would otherwise be difficult to obtain.

The dataset covers the period from 17 May 2020 to 16 May 2024, spanning approximately two and a half months. This window was chosen to coincide with the aftermath of the 28th UN Climate Change Conference (COP28, December 2023) and the subsequent months, when climate change remained a prominent topic in global public discourse.

**Search Strategy:** We employed a keyword-based search strategy to identify relevant posts. Queries were issued via the CrowdTangle API and designed to capture a wide range of climate-related narratives. The final keyword set included three categories:

- General terms: “climate change”, “global warming”, “climate crisis”, “carbon emissions”, “greenhouse gases”
- Policy and movement terms: “climate policy”, “climate action”, “climate justice”
- Activists and public figures: “Greta Thunberg” and the names of other prominent climate activists and organizations

**Metadata Collected:** For each post retrieved, we collected both page-level and post-level metadata as provided by CrowdTangle:

- P age-level: Page Name, Page Category, Page Admin Top Country, Page Description, Page Created, and audience size at posting (Likes at Posting, Followers at Posting).
- Pos t-level: textual content (Message, Image Text, Link Text, Description), temporal fields (Post Created, Date, Time), content type (Photo, Video, Link), and external reference fields (URL, Final Link).
- Engagement metrics: Likes, Comments, Shares, the full reaction taxonomy (Love, Wow, Haha, Sad, Angry, Care), Total Interactions, Overperforming Score, Post Views, and video-specific fields (e.g., Video Length, Video Share Status, Is Video Owner?).

**Data Cleaning and Integrity:** All duplicate posts were removed, and malformed entries (e.g. missing timestamps or empty messages) were removed. Given dataframe $D$ with text in *Message*, strings are coerced and empty rows removed; engagement features are numeric-coerced; timestamps are parsed when present. Minimal normalization preserves punctuation, capitalization, emojis, and hashtags.

### 2.2. Data Analysis

**Sentiment Analysis:** We quantified the sentiment characteristics of post content using polarity and subjectivity scores obtained from VADER sentiment analysis model. Polarity scores ranged from $-1$ (negative) to $+1$ (positive), while subjectivity scores ranged from $0$ (objective) to $1$ (subjective). Posts were classified into three polarity categories: Negative ($< -0.1$), Neutral ($-0.1 \leq x \leq 0.1$), and Positive ($> 0.1$). Similarly, subjectivity was grouped into three categories: Objective ($\leq 0.33$), Mixed ($0.33 < x \leq 0.67$), and Subjective ($> 0.67$).

**Engagement Classification:** To classify engagement levels, we employed quantile-based binning using the `pd.qcut()` function, creating three balanced groups—Low, Medium, and High—based on total interaction counts. Total interactions were operationalized as the cumulative sum of user reactions, including likes, comments, shares, and emotional responses associated with each post.

**Correlation Analysis:** We conducted chi-square tests of independence to examine categorical associations among sentiment variables (polarity, subjectivity), post characteristics (type, page category), and engagement levels. Contingency tables were generated using the `pd.crosstab()` function, and statistical significance was evaluated at the $\alpha = 0.05$ level. Only groups with adequate sample size ($n \geq 5$) were included in the analysis to ensure the validity of the chi-square assumptions.

| Country | Pages (K) | Page (abbr.) | Interact.(K) |
|---|---|---|---|
| US | 130 | Sky News Au | 435 |
| AU | 22 | ABC News | 333 |
| GB | 22 | CGTN | 327 |
| IN | 18 | WEF | 301 |
| CA | 14 | Fiji Govt | 251 |
| PH | 13 | Moms Clean Air | 248 |
| KE | 5.3 | Climate Reality | 235 |
| IE | 4.2 | NY Times | 234 |
| NZ | 4.1 | Reuters | 213 |
| PK | 4.0 | PBS NewsHour | 197 |

Table 1: Top countries by page and top pages by total interactions.(K) means thousands. Country name written in short form

## 3. Results

### 3.1. Dataset Summary

The dataset comprises 299,329 climate-related Facebook posts with 41 features, collected from 26,731 unique pages between May 2020 and May 2024. It includes both textual and non-textual attributes, offering a comprehensive view of user engagement and posting behavior. This four-year temporal span enables longitudinal analysis of climate communication dynamics. On average, 205 posts per day were recorded, with notable activity peaks coinciding with global events and most prominently on April 22, 2021 (Earth Day) with 1,572 posts.

Table 1 summarizes the **geographical distribution** of the pages in the dataset. The United States accounts for the largest share, with approximately 130,000 pages (around 43% of the total), followed by Australia (22K), Great Britain (22K), India (18K), and Canada (14K). Although English-speaking countries dominate the dataset, the inclusion of pages from Kenya, Pakistan, and the Philippines provides representation from the Global South, adding some international diversity.

The **leading pages** by total interactions highlight a heterogeneous mix of information sources. Sky News Australia ranks first with 435K interactions, followed by established media outlets (ABC News, The New York Times, Reuters, PBS NewsHour), intergovernmental and policy organizations (World Economic Forum), government actors (Fiji Government), and activist groups (Moms Clean Air Force, Climate Reality Project). All top ten pages registered more than 197K interactions, reflecting a high level of public engagement with climate-related content between 2020 and 2024, aligning with broader patterns of polarization in online climate discourse (Falkenberg et al., 2022).

| Post Type | Count |
|---|---|
| Link | 142,513 |
| Photo | 103,313 |
| Native Video | 38,236 |
| Live Video (Complete) | 5,633 |
| Status | 5,237 |
| Video | 1,943 |
| YouTube | 1,847 |
| Live Video (Scheduled) | 606 |
| Live Video | 1 |

Table 2: Post type distribution in the dataset

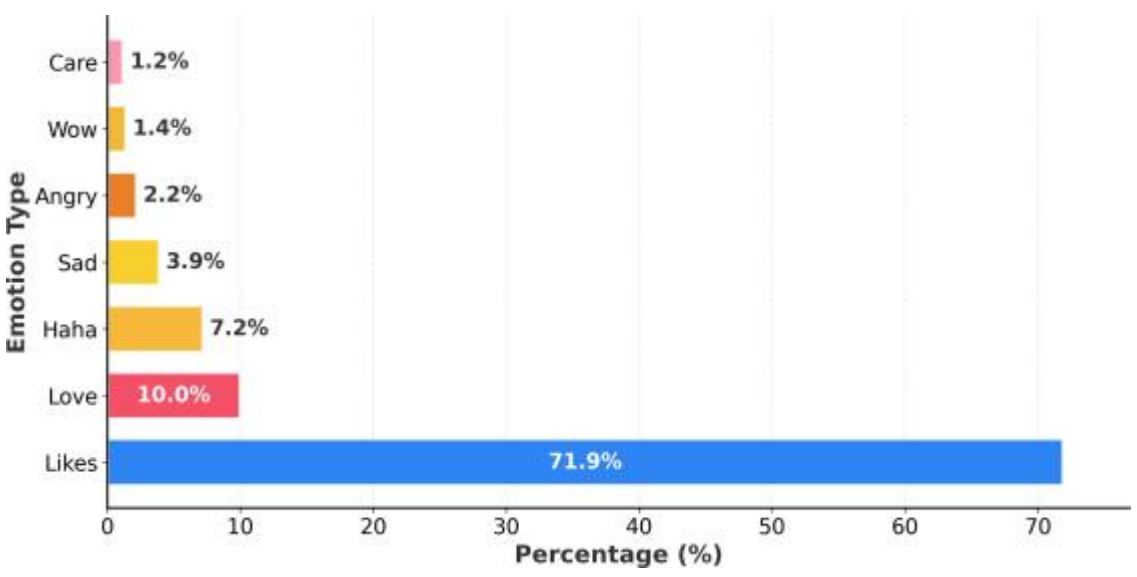

Figure 1: Emotion Reaction Distribution

Table 2 shows **post-type** climate-related Facebook content shared via links (142,513 posts), photos (103,313), and native videos (38,236), indicating a preference for visually informative formats. Less frequent use of live videos or external embeds suggests that content optimized for in-platform visibility dominates climate messaging strategies.

Figure 1 highlights that **user engagement** is largely positive or neutral, with “Likes” comprising 71.9% of reactions, followed by “Love” (10.0%) and “Haha” (7.2%). Negative emotions such as “Sad” (3.9%), “Angry” (4.2%), and “Care” (1.2%) are less common, reflecting the prevalence of passive or approving responses over contentious ones. This diversity, paired with high engagement levels, underscores Facebook’s role as a contested yet active space for global climate discourse during 2020–2024.

### 3.2. User Sentiment and Polarity

Table 3 and the accompanying chart show that objective posts overwhelmingly dominate the dataset, comprising 263,011 entries, or roughly 96.8% of total content. Subjective posts are minimal at 7138 (2.5%) in the distribution. This distribution suggests that climate-related discussions on Facebook are largely factual or report-based rather than overtly opinionated.

| Category | Subjectivity | Polarity |
|---|---|---|
| Obj/Pos | 217947 (96.82%) | 141269 (52.29%) |
| Subj/Neg | 7138 (2.51%) | 72674 (26.90%) |
| Neutral | - | 56206 (20.81%) |

Table 3: Distribution of subjectivity and polarity

| Variable | Chi-Square | P-Value |
|---|---|---|
| Page Category | 10416.84 | 0.000 |
| Type | 872.95 | 0.000 |
| Video Share Status | 352.00 | 0.000 |
| Is Video Owner? | 230.89 | 0.000 |
| Subjectivity Level | 112.93 | 0.000 |
| Polarity Sentiment | 286.92 | 0.000 |

Table 4: Chi-square tests of independence between categorical variables and engagement levels. All results are significant at $\alpha = 0.05$.

In contrast, polarity exhibits greater variation. Positive sentiment is most prevalent, with 141,269 posts (52.3%), followed by negative sentiment at 72,674 posts (26.9%), and neutral sentiment at 56,206 posts (20.8%). This balance indicates that while most posts convey optimism or supportive attitudes toward climate issues, a substantial portion still reflects skepticism, criticism, or concern. The combination of predominantly objective tones with diverse sentiment polarity suggests that much of the climate discourse on Facebook focuses on information sharing while still evoking mixed emotional responses.

### 3.3. Correlation between Total Interaction Vs Different Variables

Table 4 shows the results of Chi-square tests examining how different content features relate to engagement levels. All tested variables are significantly associated with engagement ($p < 0.05$), meaning they each play a role in how users interact with posts. The strongest effect comes from *Page Category* ($\chi^2$ = 10,416.84), suggesting that the type of page—such as media, government, or advocacy—has a major influence on engagement. *Post Type* ($\chi^2$ = 872.95) and *Video Share Status* ($\chi^2$ = 352.00) also have strong effects, showing that the format and sharing method of a post matter. Sentiment-related features like *Polarity Sentiment* ($\chi^2$ = 286.92) and *Subjectivity Level* ($\chi^2$ = 112.93) also affect engagement, indicating that emotional tone shapes how people respond. Overall, both the structure and emotional content of posts are important factors in driving interaction around climate issues on Facebook.

### 3.4. Topic Modeling and Thematic Analysis

**Thematic Analysis:** To explore the main thematic structures within climate-related Facebook discourse, we applied *Latent Dirichlet Allocation (LDA)* to the dataset. From the resulting topic–word distributions, we extracted the most frequent keywords and manually assigned topic labels using domain expertise and contextual interpretation. The process produced top ten coherent topic. Finally, we applied radial framework to find interconnected topic and themes (see Figure 2). The radial framework (Bachmaier, 2007) presents a comprehensive organizational structure of climate and environmental research, encompassing 10 distinct topics distributed across 5 thematic areas.

- **Policy & Governance:** Includes *U.S. Climate Policy*, *India's Environmental Governance*, and *Climate Agreements*. Highlights the central role of political institutions and leadership in enabling climate action at both national and global scales.
- **Activism & Public Engagement:** Comprises *Youth Activism* and *Public Protests*, showcasing how grassroots mobilization—especially by youth—drives urgency, accountability, and public awareness in climate debates.
- **International Cooperation:** Represented by *Climate Summits & Global Leadership*. Captures the global dimension of climate action through platforms like COP meetings that shape collective policy commitments.
- **Science & Technology:** Covers *Emissions Reduction & Clean Energy* and *Extreme Weather & Climate Science*, bridging observational insights with technological solutions to mitigate and monitor climate risks.
- **Conservation & Development:** Encompasses *Biodiversity Conservation* and *Sustainable Development*, linking environmental protection with equitable growth and recognizing the co-dependence of ecological and human systems.

**User Engagement in Topic Analysis:** Topics emphasizing public action and policy, including *Environmental Activism and Public Protests* and *Climate Policy and Global Agreements*, yielded the highest average total interactions (218.11 and 217.93, respectively). These topics also recorded elevated commenting activity, reflecting their potential to stimulate discussion and debate. Such findings indicate strong deliberative or polarizing potential.

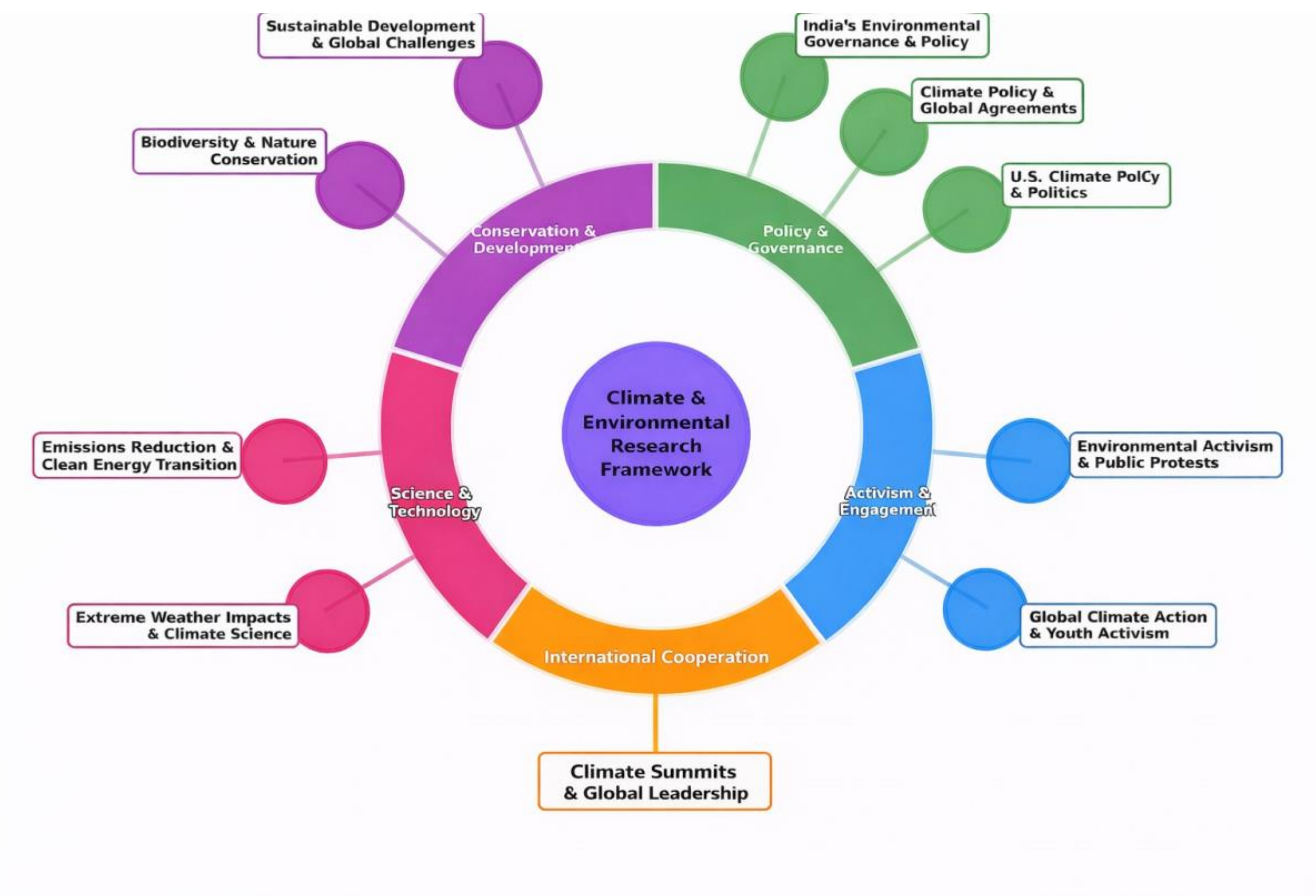

Figure 2: Climate and environmental research topics organized in a radial thematic framework. 10 Research Topic, 5 Thematic Areas interconnected framework

Interestingly, themes such as *International Climate Summits and Global Leadership* and *Global Climate Action and Youth Activism* achieved exceptional reach, as evidenced by the highest average post view counts indicating broad audience reach—likely driven by international media cycles particularly around major UN events and youth-led movements.

In contrast, topics such as *India's Environmental Governance* demonstrated relatively high "like" counts but lower levels of deeper engagement (e.g., shares, comments), possibly reflecting geographically localized relevance or algorithmic audience segmentation.

Overall, the findings point to a complex communication landscape in which emotionally salient, politically charged, and visually oriented topics elicit distinct patterns of public engagement. By integrating unsupervised topic modeling with behavioral metrics, this analysis highlights the different ways users express support, interest, or critique across themes.

Both the dominant narratives driving online climate discourse and the ways in which different thematic frames shape public attention and interaction. These insights are valuable for researchers and communicators seeking to understand or influence the evolving dynamics of climate communication in networked media environments.

## 4. Discussions

### 4.1. Evaluation and Data Quality

The whole dataset was annotated by unsupervised VADER sentiment analysis model. To evaluate the performance of unsupervised model, we randomly selected 1000 samples and annotated it. Then, we compare the manually annotated data with the sampled data produced by VADER model. Table 6 shows the evaluation These results indicate that while VADER performs reasonably well in identifying subjectivity, it struggles with the nuance of sentiment detection, especially when the sentiment distribution is heavily skewed toward the "Positive" class. These disparities highlight that domain-specific annotation improves both agreement and robustness, and that noisy, off-the-shelf tools without adaptation to climate discourse risk introducing bias. Overall, data quality—ensured through careful manual annotation—proved decisive in benchmarking and guiding model selection.

### 4.2. Dataset Release

The climatechat-300K dataset presented in this study provides a unique and comprehensive re-source for computational social science, communication studies, and climate policy research. With more than 299,329 posts, spanning over four years and enriched by 41 metadata attributes, it enables scholars to move beyond surface-level sentiment and engagement metrics toward a deeper understanding of how climate discourse evolves across time, geographies, and institutional actors.

| Topic Name | Total Interactions | Likes | Comments | Shares | Views |
|---|---|---|---|---|---|
| U.S. Climate Policy and Politics | 214.96 | 86.50 | 58.81 | 13.57 | 504.57 |
| Sustainable Development and Global Challenges | 156.52 | 110.00 | 13.26 | 12.57 | 1015.35 |
| Global Climate Action and Youth Activism | 187.29 | 104.21 | 30.74 | 16.49 | 1658.53 |
| International Climate Summits and Global Leadership | 190.97 | 105.12 | 33.25 | 15.34 | 2003.57 |
| Emissions Reduction and Clean Energy Transition | 175.55 | 92.15 | 35.23 | 15.35 | 1195.64 |
| Environmental Activism and Public Protests | 218.11 | 86.67 | 66.02 | 12.50 | 1423.09 |
| Climate Policy and Global Agreements | 217.93 | 101.77 | 51.11 | 10.82 | 742.16 |
| Extreme Weather Impacts and Climate Science Reports | 177.63 | 72.46 | 37.23 | 17.87 | 963.57 |
| Biodiversity and Nature Conservation | 167.68 | 105.30 | 13.47 | 18.57 | 1349.39 |
| India's Environmental Governance and Climate Policy | 128.48 | 106.40 | 6.14 | 8.94 | 973.22 |

Table 5: Engagement metrics by topic cluster. Values represent average interactions per post.

| Task | Accu | Prec | Rec | F1 |
|---|---|---|---|---|
| Sentiment | 0.61 | 0.45 | 0.43 | 0.42 |
| Subjectivity | 0.78 | 0.68 | 0.61 | 0.64 |

Table 6: Evaluation Metrics (Macro Avg)

Our analyses demonstrate the potential of this dataset for multiple research directions. Topic modeling highlights dominant narratives such as climate policy, renewable energy, and environmental conservation, while stance and sentiment analysis (Biswas and Zaghouani, 2025) reveal how polarization, emotional intensity, and subjectivity shape public engagement. Metadata dimensions including page categories, geographic origin, and post formats further allow researchers to link content strategies with interaction patterns, offering insight into how media, governments, and advocacy groups frame climate change.

Importantly, the **planned release** of this dataset to the research community represents a critical step toward advancing transparency and reproducibility in computational social science. Publicly available large-scale social media datasets remain scarce due to access restrictions and ethical considerations (Souza et al., 2014). By providing this resource, we enable scholars from diverse disciplines—including computer science, linguistics, sociology, political science, and public health—to pursue investigations into misinformation, public opinion formation, cultural discourse, and the diffusion of ideas.

### 4.3. Future Research Direction

Several promising avenues emerge from this work. The dataset offers opportunities for cross-domain and multilingual analysis, particularly in assessing the generalizability of models across different cultural and linguistic contexts (Ruder et al., 2019; Bingler et al., 2023). Then, the longitudinal nature of the data enables temporal studies of evolving discourse patterns, which can shed light on the dynamics of polarization, opinion shifts, and responses to major societal events (Yuan et al., 2013). Finally, this dataset provides an important platform for advancing responsible AI (Baeza-Yates, 2024) research, where fairness, interpretability, and ethical considerations are paramount in the development of computational models for sensitive social phenomena.

### 4.4. Broader Societal Impact

**Positive Impacts.** The *ClimateChat-300K* dataset supports researchers, journalists, and policymakers seeking to understand global climate communication on social media. By providing large-scale and systematically collected data, it enables more transparent and reproducible studies of how online engagement shapes environmental awareness and discourse. The resource contributes to equity and inclusion by incorporating perspectives from multiple geographic regions, including voices from both the Global North and the Global South. It helps expand research beyond well-studied platforms and offers opportunities to design more inclusive climate communication strategies.

**Potential Risks.** Like all social media datasets, this resource carries the possibility of unintended misuse. Analyses derived from it could be employed to monitor or profile specific communities, reinforce political polarization, or misrepresent public sentiment about climate policy. Since the dataset is limited to publicly available material, representation biases and algorithmic amplification patterns may also influence findings, potentially leading to incomplete or skewed interpretations.

**Mitigation and Recommendations.** The dataset will be distributed for academic and non-commercial research only, under a license that requires responsible use and proper attribution. Users will be advised to exercise caution when drawing inferences about individuals or communities. Detailed documentation will describe data collection boundaries, ethical safeguards, and recommended citation practices. We encourage collaboration with climate communication experts and affected communities to promote fair and context-aware interpretations of the data.

## 5. Dataset Availability and License

The **ClimateChat-300K dataset** is publicly available via Zenodo[1]. Access to the dataset requires completion of a request form[2].

The dataset is released strictly for research purposes. All materials are distributed under the *Creative Commons Attribution–NonCommercial–ShareAlike 4.0 International License (CC BY-NC-SA 4.0)*. Users must comply with the terms of this license, including proper attribution, non-commercial use, and distribution of derivative works under the same license.

## 6. Conclusion

*ClimateChat-300K* represents one of the most comprehensive open resources to date for analyzing global climate communication on Facebook. By compiling nearly three hundred thousand posts from more than twenty-six thousand pages, the dataset provides a unique opportunity to study the complex interplay between scientific information, public discourse, and digital engagement across multiple years of global climate debate. Its metadata-rich design, covering forty-one contextual and behavioral features, allows researchers to explore questions that extend beyond sentiment or popularity, including how different actors, regions, and communication styles shape online conversations about climate change.

Through the integration of large-scale text analysis, sentiment assessment, and topic modeling, our initial findings reveal clear relationships between emotional tone, content type, and audience interaction. These insights emphasize the importance of communication framing and platform affordances in influencing public attention and engagement. Beyond descriptive analyses, the dataset can serve as a foundation for predictive modeling, comparative studies across platforms, and interdisciplinary collaborations bridging linguistics, social science, and environmental communication.

Looking ahead, we aim to expand this resource by incorporating multilingual and multimodal data to capture broader global perspectives on climate issues. We also encourage the research community to contribute feedback, methodological improvements, and additional annotations to enhance future releases. By releasing *ClimateChat-300K*, we hope to foster transparent, inclusive, and responsible research that advances both computational social science and the public understanding of climate communication in the digital age.

## 7. Limitations

While the *ClimateChat-300K* dataset represents a substantial effort to document global online discourse on climate change, several limitations must be acknowledged.

First, the dataset exclusively covers publicly available Facebook Pages, which may not fully reflect broader climate-related conversations occurring on other social media platforms, private groups, or offline spaces. This focus introduces potential platform-specific and demographic biases, since Facebook audiences differ in age, geography, and engagement style compared to users of platforms such as X or TikTok.

Second, data collection relied on a keyword-based approach that may omit relevant discussions framed in less explicit or alternative language. Despite efforts to design comprehensive search queries, this method inevitably introduces topical and linguistic selection bias.

Third, the dataset captures a fixed four-year period between 2020 and 2024, reflecting a historical snapshot rather than a continuously updated resource. The closure of the CrowdTangle API in 2024 further restricts replicability and ongoing data acquisition.

Fourth, while automated sentiment analysis and topic modeling provide valuable insights, they may misclassify nuanced expressions, sarcasm, or culturally contextual language. Manual validation was limited to a small sample size, and inter-annotator agreement could not be fully quantified.

Finally, engagement metrics such as likes, shares, and comments serve as indirect indicators of attention rather than verified measures of influence or attitudinal change. Researchers are therefore cautioned against inferring causality from observed patterns.

Future work will focus on expanding coverage to additional languages and platforms, refining annotation quality, and incorporating multimodal and temporal dynamics to better capture the evolving landscape of climate communication.

[1] https://zenodo.org/records/18824386
[2] https://forms.gle/W7xpLt7io326bR3A6

## 8. Ethics Statement

This work adheres to established ethical standards in computational social science and linguistic data research. All data were obtained exclusively from publicly accessible Facebook Pages using the official CrowdTangle API prior to its retirement in 2024. No private or restricted content was accessed, and no attempts were made to infer personal identities or link behavioral data to individual users.

To protect privacy, all personally identifiable information (PII) was removed or anonymized during preprocessing. Page-level metadata were included only when publicly available and relevant to the analysis. The dataset therefore contains no individual user identifiers, direct messages, or comments from private groups.

The dataset complies with Facebook's data use and platform terms of service, as well as with general principles of data minimization and fairness under frameworks such as the GDPR and the ACM Code of Ethics. Researchers are encouraged to review applicable data protection laws and institutional ethical review requirements prior to using the dataset.

Potential risks include the possibility that derived analyses could be used to profile or target specific communities, or that aggregated patterns might inadvertently reinforce existing political or regional biases. To mitigate these risks, access will be restricted to academic and non-commercial research, with accompanying documentation that emphasizes responsible and context-aware use.

Finally, this project recognizes the importance of equitable climate communication. The dataset is intended to advance transparency, reproducibility, and inclusivity in climate discourse research while avoiding harm to affected or underrepresented populations.

## Acknowledgments

This research was made possible by the National Priorities Research Program grant NPRP14C-0916-210015 from the Qatar National Research Fund (QNRF), part of the Qatar Research, Development and Innovation Council (QRDI).

## 9. References

Christian Bachmaier. 2007. A radial adaptation of the sugiyama framework for visualizing hierarchical information. *IEEE Transactions on Visualization and Computer Graphics*, 13(3):583–594.

Ricardo Baeza-Yates. 2024. Introduction to responsible ai. In *Proceedings of the 17th ACM International Conference on Web Search and Data Mining*, pages 1114–1117.

Julia Bingler, Mathias Kraus, Markus Leippold, and Nicolas Webersinke. 2023. How cheap talk in climate disclosures relates to climate initiatives, corporate emissions, and reputation risk. Working paper, Available at SSRN 3998435.

Md. Rafiul Biswas and Wajdi Zaghouani. 2025. Enhancing Arabic dialectal sentiment analysis through advanced data augmentation techniques. In *Proceedings of the Shared Task on Sentiment Analysis for Arabic Dialects*, pages 24–28, Varna, Bulgaria. INCOMA Ltd., Shoumen, Bulgaria.

Meta Transparency Center. 2024. Crowdtangle. `https://transparency.meta.com/he-il/researchtools/other-datasets/crowdtangle/`. Accessed: 2025-09-11.

Kirtika Deo and Abhnil Amtesh Prasad. 2020. Evidence of climate change engagement behaviour on a facebook fan-based page. *Sustainability*, 12(17):7038.

Max Falkenberg, Alessandro Galeazzi, Maddalena Torricelli, Niccolò Di Marco, Francesca Larosa, Madalina Sas, Amin Mekacher, Warren Pearce, Fabiana Zollo, Walter Quattrociocchi, et al. 2022. Growing polarization around climate change on social media. *Nature Climate Change*, 12(12):1114–1121.

Aleksandrina V Mavrodieva, Okky K Rachman, Vito B Harahap, and Rajib Shaw. 2019. Role of social media as a soft power tool in raising public awareness and engagement in addressing climate change. *Climate*, 7(10):122.

Niels G Mede and Ralph Schroeder. 2024. The "greta effect" on social media: A systematic review of research on thunberg's impact on digital climate change communication. *Environmental Communication*, 18(6):801–818.

Angelina Mooseder, Cornelia Brantner, Rodrigo Zamith, and Jürgen Pfeffer. 2023. (social) media logics and visualizing climate change: 10 years of# climatechange images on twitter. *Social Media+ Society*, 9(1):20563051231164310.

Sebastian Ruder, Anders Søgaard, and Ivan Vulić. 2019. Unsupervised cross-lingual representation learning. In *Proceedings of the 57th Annual Meeting of the Association for Computational Linguistics: Tutorial Abstracts*, pages 31–38.

Alexandra Segerberg. 2017. Online and social media campaigns for climate change engagement.

In *Oxford research encyclopedia of climate science*. Oxford University Press.

Giuliana Sorce and Delia Dumitrica. 2023. From school strikes to webinars: Mapping the forced digitalization of fridays for future's activism during the covid-19 pandemic. *Convergence*, 29(3):570–585.

Roberto CSNP Souza, Denise EF de Brito, Rodrigo L Cardoso, Derick M de Oliveira, Wagner Meira Jr, and Gisele L Pappa. 2014. An evolutionary methodology for handling data scarcity and noise in monitoring real events from social media data. In *Ibero-American Conference on Artificial Intelligence*, pages 295–306. Springer.

Saverio Storani, Max Falkenberg, Walter Quattrociocchi, and Matteo Cinelli. 2025. Relative engagement with sources of climate misinformation is growing across social media platforms. *Scientific Reports*, 15(1):18629.

Bebe Chand Sultana, Md Tabiur Rahman Prodhan, Edris Alam, Md Salman Sohel, ABM Mainul Bari, Subodh Chandra Pal, Md Kamrul Islam, and Abu Reza Md Towfiqul Islam. 2024. A systematic review of the nexus between climate change and social media: present status, trends, and future challenges. *Frontiers in Communication*, 9:1301400.

Quan Yuan, Gao Cong, Zongyang Ma, Aixin Sun, and Nadia Magnenat Thalmann. 2013. Who, where, when and what: discover spatio-temporal topics for twitter users. In *Proceedings of the 19th ACM SIGKDD international conference on Knowledge discovery and data mining*, pages 605–613.